\documentclass[10pt,twocolumn,letterpaper]{article}

\usepackage{cvpr}
\usepackage[utf8]{inputenc}
\usepackage{times}
\usepackage{epsfig}
\usepackage{graphicx}
\usepackage{amsmath}
\usepackage{amssymb}
\usepackage{comment}
\usepackage{tikz}

\definecolor{gold}{HTML}{BD820B}
\definecolor{silver}{HTML}{909090}
\definecolor{bronze}{HTML}{9A5F26}
\newcommand*\circledd[1]{\tikz[baseline=(char.base)]{
            \node[shape=circle,draw,inner sep=0.15pt] (char) {#1};}}           
\newcommand{\first}[1]{%
    {\raisebox{0.8pt}{\footnotesize \color{gold} \circledd{1}}\hspace{3.5pt}{\bf #1}}%
}
\newcommand{\second}[1]{%
    {\raisebox{0.8pt}{\footnotesize \color{silver} \circledd{2}}\hspace{3.5pt}#1}%
}
\newcommand{\third}[1]{%
    {\raisebox{0.8pt}{\footnotesize \color{bronze} \circledd{3}}\hspace{3.5pt}#1}%
}

\graphicspath{{./figures/}}

\usepackage[pagebackref=true,breaklinks=true,letterpaper=true,colorlinks,bookmarks=false]{hyperref}

\cvprfinalcopy 

\def\cvprPaperID{7249} 

\ifcvprfinal\fi

\begin{document}

\title{D3S -- A Discriminative Single Shot Segmentation Tracker}

\author{\vspace{-1cm} \\ Alan Lukežič$^1$, Jiří Matas$^2$, Matej Kristan$^1$\\
{\small $^1$Faculty of Computer and Information Science, University of Ljubljana, Slovenia} \\
{\small $^2$Faculty of Electrical Engineering, Czech Technical University in Prague, Czech Republic} \\
{\tt\small alan.lukezic@fri.uni-lj.si}
\vspace{-0.5cm}
}

\maketitle
\thispagestyle{empty}

\begin{abstract}
Template-based discriminative trackers are currently the dominant tracking paradigm due to their robustness, but are restricted to bounding box tracking and a limited range of transformation models, which reduces their localization accuracy.
We propose a discriminative single-shot segmentation tracker -- D3S, which narrows the gap between visual object tracking and video object segmentation. 
A single-shot network applies two target models with complementary geometric properties, one invariant to a broad range of transformations, including non-rigid deformations, the other assuming a rigid object to simultaneously achieve high robustness and online target segmentation.
Without per-dataset finetuning and trained only for segmentation as the primary output, D3S outperforms all trackers on VOT2016, VOT2018 and GOT-10k benchmarks and performs close to the  state-of-the-art trackers on the TrackingNet. 
D3S outperforms the leading segmentation tracker SiamMask on video  object segmentation benchmarks and performs on par with top video object segmentation algorithms, while running an order of magnitude faster, close to real-time.
PyTorch implementation is available here: \url{https://github.com/alanlukezic/d3s}
\end{abstract}

\section{Introduction}  \label{sec:introduction}

Visual object tracking is one of core computer vision problems. 
The most common formulation considers the task of reporting target location in each frame of the video given a single training image. Currently, the dominant tracking paradigm, performing best in evaluations~\cite{kristan_vot2017,kristan_vot2018}, 
is correlation bounding box tracking~\cite{DanelljanCVPR2017, danelljan_eccv2018_updt, Lukezic_CVPR_2017,siamfc_eccv16,dasiamrpn_eccv2018,siamrpn_cvpr2018} where the target represented by a  multi-channel rectangular template 
is localized by cross-correlation between the template and a search region.

\begin{figure}[!t]
\centering
\includegraphics[width=1\linewidth]{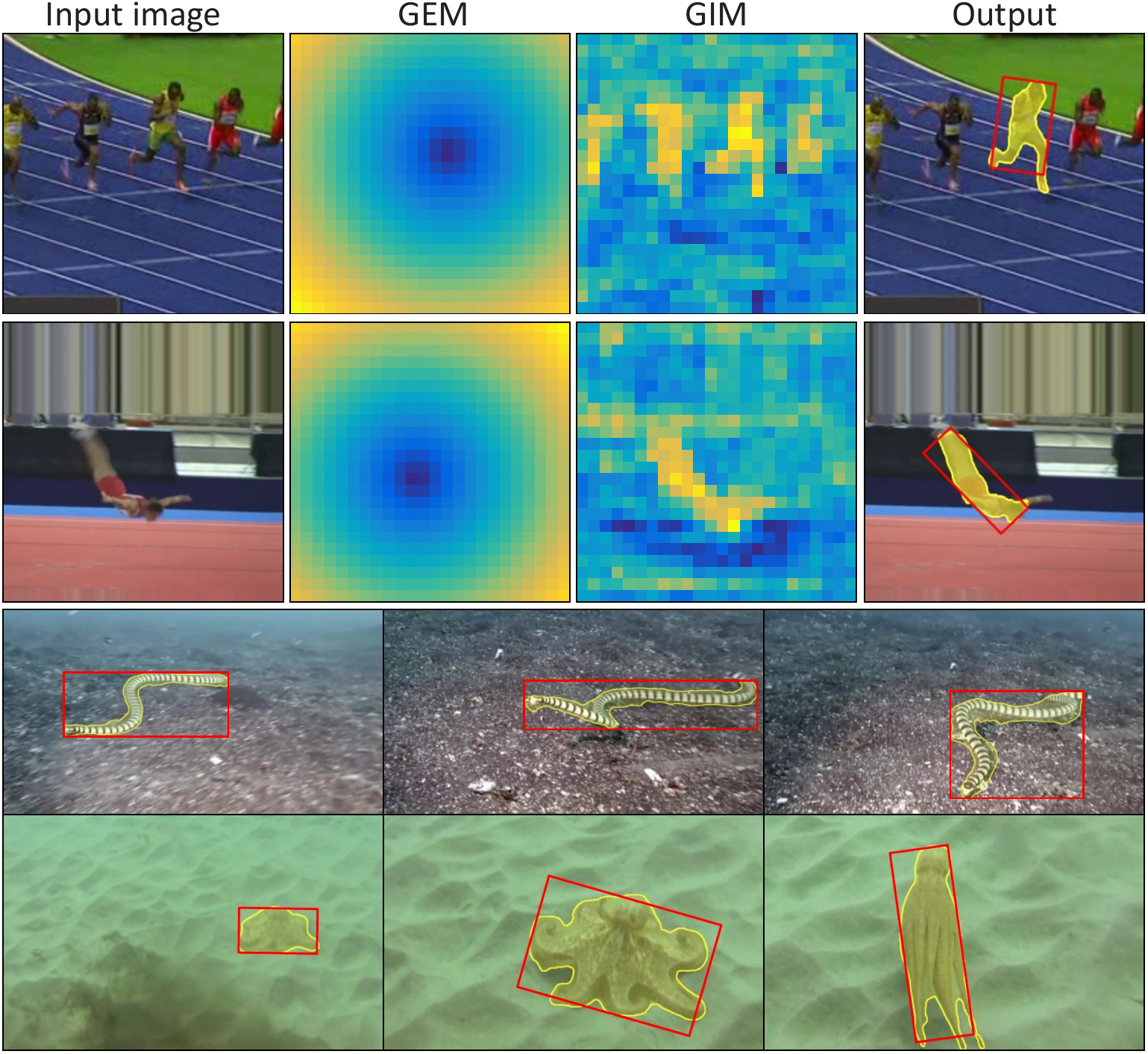}
\caption{
The D3S tracker represents the target by two models with complementary geometric properties, one invariant to a broad range of transformations, including non-rigid deformations (GIM - geometrically invariant model), the other assuming a rigid object with motion well approximated by an euclidean transformation (GEM - geometrically constrained Euclidean model). 
The D3S, exploiting the complementary strengths of GIM and GEM, provides both state-of-the-art localisation and accurate segmentation, even in the presence of substantial deformation.
}
\label{fig:intro_figure}
\end{figure}  

State-of-the-art template-based trackers apply an efficient brute-force search for target localization. 
Such strategy is appropriate for low-dimensional transformations like translation and scale change, but becomes inefficient for more general situations e.g. such that induce an aspect ratio change and rotation. 
As a compromise, modern trackers combine approximate exhaustive search with sampling and/or bounding box refinement/regression networks~\cite{atom_cvpr19,siamrpn_cvpr2019} for aspect ratio estimation.
However, these approaches are restricted to axis-aligned rectangles. 

Estimation of high-dimensional template-based transformation is unreliable when a bounding box is a poor approximation of the target~\cite{lukezic_dpt}.
This is common --  consider e.g. elongated, rotating, deformable objects, or a person with spread out hands. 
In these cases, the most accurate and well-defined target location model is a binary per-pixel segmentation mask.
If such output is required, tracking becomes the video object segmentation task recently popularized by DAVIS~\cite{davis16,davis17} and YoutubeVOS~\cite{yt_vos2018} challenges. 

Unlike in tracking, video object segmentation challenges typically consider large targets observed for less than 100 frames with low background distractor presence. 
Top video object segmentation approaches thus fare poorly in short-term tracking scenarios~\cite{kristan_vot2018} where the target covers a fraction of the image, substantially changes its appearance over a longer period and moves over a cluttered background. 
Best trackers apply visual model adaptation, but in the case of segmentation errors it leads to an irrecoverable tracking failure~\cite{dat_cvpr15}. 
Because of this, in the past, segmentation has played only an auxiliary role in template-based trackers~\cite{staple_cvpr2016}, constrained DCF learning~\cite{Lukezic_CVPR_2017} and tracking by 3D model construction~\cite{otr_cvpr19}.

Recently, the SiamRPN~\cite{siamrpn_cvpr2018} tracker has been extended to produce high-quality segmentation masks in two stages~\cite{siammask_cvpr19} -- the target bounding box is first localized by SiamRPN branches and then a segmentation mask is computed only within this region by another branch. 
The two-stage processing misses the opportunity to treat localization and segmentation jointly to increase robustness. 
Another drawback is that a fixed template is used that
cannot be discriminatively adapted to the changing scene.

We propose a new single-shot discriminative segmentation tracker, D3S, that addresses the above-mentioned limitations. 
The target is encoded by two discriminative visual models -- one is adaptive and highly discriminative, but geometrically constrained to an Euclidean motion (GEM), while the other is invariant to broad range of transformation (GIM, geometrically invariant model), see Figure~\ref{fig:intro_figure}. 

GIM sacrifices spatial relations to allow target localization under significant deformation. On the other hand, GEM predicts only position, but discriminatively adapts to the target and acts as a selector between possibly multiple target segmentations inferred by GIM. 
In contrast to related trackers~\cite{siammask_cvpr19,siamrpn_cvpr2019,atom_cvpr19}, the primary output of D3S is a segmentation map computed in a single pass through the network, which is trained end-to-end for segmentation only (Figure~\ref{fig:architecture-overview}). 

Some applications and most tracking benchmarks require reporting the target location as a bounding box. 
As a secondary contribution, we propose an effective method for interpreting the segmentation mask as a rotated rectangle. This avoids an error-prone greedy search and naturally addresses changes in location, scale, aspect ratio and rotation.

D3S outperforms all state-of-the-art trackers on most of the major tracking benchmarks~\cite{kristan_vot2016,kristan_vot2018,got10k,muller_trackingnet} despite not being trained for bounding box tracking. 
In video object segmentation benchmarks~\cite{davis16,davis17}, D3S outperforms the leading segmentation tracker~\cite{siammask_cvpr19} and performs on par with top video object segmentation algorithms (often tuned to a specific domain), yet running orders of magnitude faster. 
Note that D3S is not re-trained for different benchmarks -- 
a single pre-trained version shows remarkable generalization ability and versatility.

\section{Related Work}  \label{sec:related_work}

Robust localization crucially depends on the discrimination capability between the target and the background distractors. 
This property has been studied in depth in discriminative template trackers called discriminative correlation filters (DCF)~\cite{bolme2010visual}. The template learning is formulated as a (possibly nonlinear) ridge regression problem and solved by circular correlation~\cite{bolme2010visual,danelljan2014adaptive,henriques2015tracking,samf_eccv2014}. 
While trackers based purely on color segmentation~\cite{comanichu_kernel_pami2003,dat_cvpr15} are inferior to DCFs, segmentation has been used for improved DCF tracking of non-rectangular targets~\cite{staple_cvpr2016,lukezic_dpt}. 
Lukežič et al.~\cite{Lukezic_CVPR_2017} used color segmentation to constrain DCF learning and proposed a real-time tracker with hand-crafted features which achieved performance comparable to trackers with deep features. 
The method was extended to long-term~\cite{Lukezic_ACCV_2018} and RGB-depth tracking~\cite{otr_cvpr19} using color and depth segmentation. 
Further improvements in DCF tracking considered deep features: Danelljan et al.~\cite{DanelljanCVPR2017} used features pre-trained for detection, Valmadre et al.~\cite{Valmadre_2017_CVPR} proposed pre-training features for DCF localization and recently Danelljan et al.~\cite{atom_cvpr19} proposed a deep DCF training using backpropagation.

Another class of trackers, called Siamese trackers~\cite{siamfc_eccv16,tao2016sint,sa_siam_cvpr2018}, has evolved in direction of generative templates. 
Siamese trackers apply a backbone pre-trained offline with general targets such that object-background discrimination is maximized by correlation between the search region and target template extracted in the first frame~\cite{siamfc_eccv16}.
The template and the backbone are fixed during tracking, leading to an excellent real-time performance~\cite{kristan_vot2018}. 
Several multi-stage Siamese extensions have been proposed. These include addition of region proposal networks for improved target localization accuracy~\cite{siamrpn_cvpr2018,siamrpn_cvpr2019} and addition of segmentation branches~\cite{siammask_cvpr19} for accurate target segmentation. 
Recently a template adaptation technique by backprop has been proposed~\cite{gradnet_iccv2019} to improve tracking robustness.

Segmentation of moving objects is a central problem in the emerging field of video object segmentation (VOS)~\cite{davis16,yt_vos2018}. 
Most recent works~\cite{feelvos_cvpr2019,osvos_cvpr2017,onavos_bmvc2017,favos_cvpr2018,osmn_cvpr2018} achieve impressive results, but involve large deep networks, which often require finetuning and are slow.
Hu et al.~\cite{videomatch_eccv2018} and Chen et al.~\cite{blazingly_fast_cvpr18} concurrently proposed segmentation by matching features extracted in the first frame, which considerably reduces the processing time. 
However, the VOS task considers segmentation of large objects with limited appearance changes in short videos. Thus, these methods fare poorly on the visual object tracking task with small, fast moving objects. 
The work proposed in this paper aims at narrowing the gap between visual object tracking and video object segmentation.

\section{Discriminative segmentation network}  \label{sec:methods}

\begin{figure}[!t]
\begin{center}
	\includegraphics[width=\linewidth]{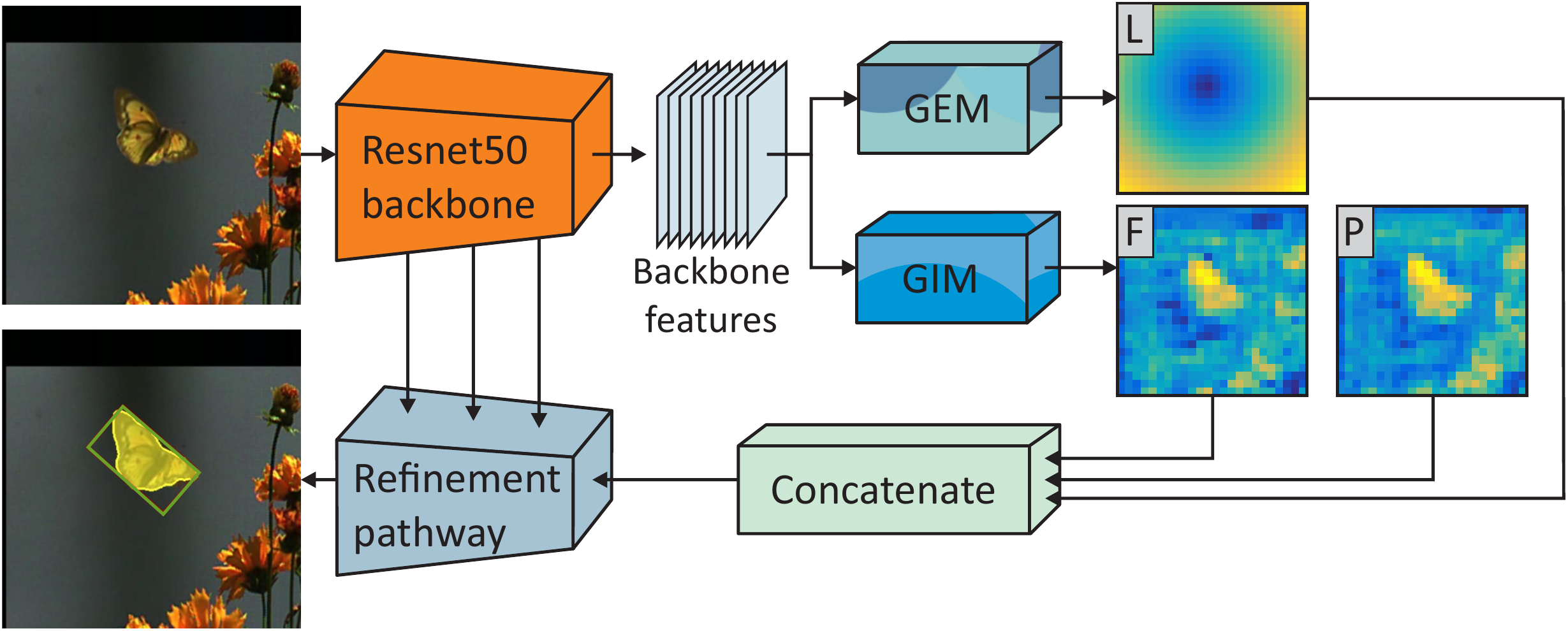}
\end{center}
   \caption{The D3S segmentation architecture. The backbone features are  processed by the GEM and GIM pathways,
   producing the target location ($\mathbf{L}$), foreground similarity ($\mathbf{F}$) and target posterior ($\mathbf{P}$) channels.
   The output of the three channels are concatenated and refined into a detailed segmentation map.} 
\label{fig:architecture-overview} 
\end{figure}

Two models are used in D3S to robustly cope with target appearance changes and background discrimination: a geometrically invariant model (GIM) presented in Section~\ref{sec:sum}, and a
geometrically constrained Euclidean model (GEM) presented in Section~\ref{sec:scm}. 
These models process the input in parallel pathways and produce several coarse target presence channels, which are fused into a detailed segmentation map by a refinement pathway described in Section~\ref{sec:refinement}. 
See Figure~\ref{fig:architecture-overview} for the architecture outline.

\subsection{Geometrically invariant model pathway}  \label{sec:sum}

Accurate segmentation of a deformable target requires loose spatial constraints in the discriminative model. 
Our geometrically invariant model (GIM) is thus composed of two 
sets of deep feature vectors corresponding to the target and the background, i.e., $\mathbf{X}_\mathrm{GIM} = \{ \mathbf{X}^{F}, \mathbf{X}^{B}\}$. 

Since the pre-trained backbone features are sub-optimal for accurate segmentation, these are first processed by a $1\times 1$ convolutional layer to reduce their dimensionality to 64, which is followed by a $3 \times 3$ convolutional layer (a ReLU is placed after each convolutional layer). 
Both these layers are adjusted in the network training stage to produce optimal features for segmentation. 
The target/background models are created in the first frame by extracting the segmentation feature vectors at pixel locations corresponding to the target ($\mathbf{X}^{F}$) and from the immediate neighborhood for the background ($\mathbf{X}^{B}$). 

\begin{figure}[!t]
\begin{center}
	\includegraphics[width=\linewidth]{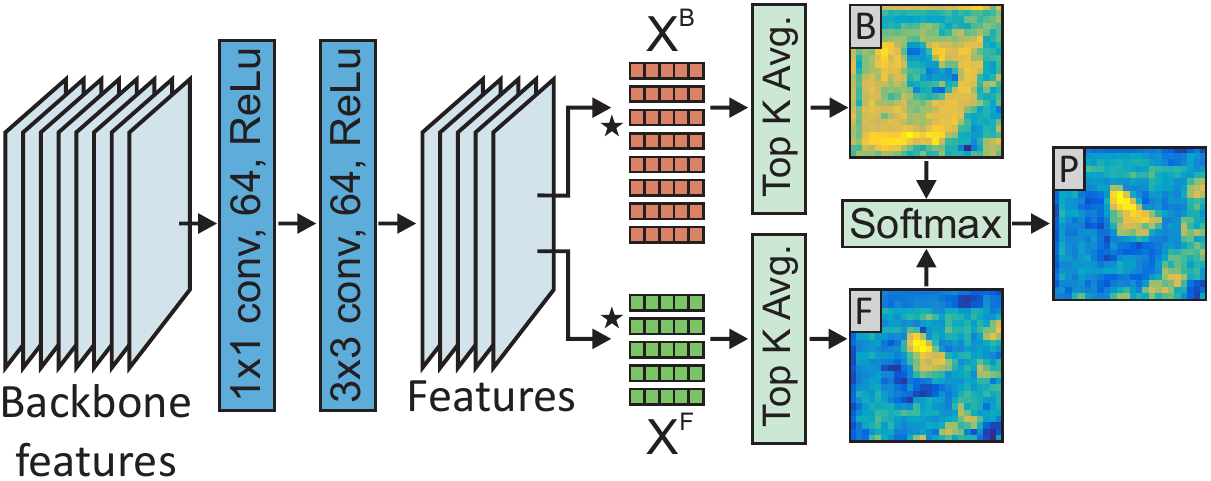}
\end{center}
   \caption{GIM -- the geometrically invariant model -- features are matched to the features in the foreground-background model $\{ \mathbf{X}^{F}, \mathbf{X}^{B}\}$ to obtain the target ($\mathbf{F}$) and background ($\mathbf{B}$) similarity channels.
   The posterior channel ($\mathbf{P}$) is the  softmax of $\mathbf{F}$ and $\mathbf{B}$.} 
\label{fig:sum-architecture} 
\end{figure}

During tracking, the pixel-level features extracted from the search region are compared to those of GIM ($\mathbf{X}_\mathrm{GIM}$) to compute foreground and background similarity channels $\mathbf{F}$ and $\mathbf{B}$ following~\cite{videomatch_eccv2018}. 
Specifically, for the $\mathbf{F}$ channel computation, each feature $\mathbf{y}_i$ extracted at pixel $i$ is compared to all features $\mathbf{x}_j^F \in \mathbf{X}^{F}$ by a normalized dot product
\begin{equation}  \label{eq:sum-similarity}
    s_{ij}^{F}(\mathbf{y}_i, \mathbf{x}_j^F) = \langle \tilde{\mathbf{y}}_i, \tilde{\mathbf{x}}_j^F \rangle,
\end{equation}
where $\tilde{(\cdot )}$ indicates an $L_2$ normalization. The final per-pixel foreground similarity at pixel $i$, $\mathbf{F}_i$, is obtained by average of top-K similarities at that pixel, i.e., 
\begin{equation}  \label{eq:fg-similarity}
    \mathbf{F}_i = \mathrm{TOP}( \{s_{ij}^{F}\}_{j=1:N_F}, K ),
\end{equation}
where $\mathrm{TOP}(\cdot, K)$ is a top-K averaging operator over the set of $N_F$ similarities. 
Computation of the background similarity channel $\mathbf{B}$ follows the same principle, but with similarities computed with the background model feature vectors, i.e., $\mathbf{x}_j^B \in \mathbf{X}^{B}$. Finally, a softmax layer is applied to produce a target posterior channel $\mathbf{P}$.
The GIM pathway architecture is shown in Figure~\ref{fig:sum-architecture}.

\subsection{Geometrically constrained model pathway}  \label{sec:scm}

While GIM produces an excellent target-background separation, it cannot well distinguish the target from similar instances, leading to a reduced robustness (see Figure~\ref{fig:intro_figure}, first line). 
Robust localization, however, is a well-established quality of the discriminative correlation filters. 
Although they represent the target by a geometrically constrained model (i.e., a rectangular filter), efficient techniques developed to adapt to the target discriminative features~\cite{danelljan_eccv2016_ccot,Lukezic_CVPR_2017,atom_cvpr19} allow tracking reliably under considerable appearance changes.

We thus employ a recent deep DCF formulation~\cite{atom_cvpr19} in the geometrically constrained Euclidean model (GEM) pathway. 
Following~\cite{atom_cvpr19}, the backbone features are first reduced to 64 channels by $1 \times 1$ convolutional layer. 
The reduced features are correlated by a 64 channel DCF followed by a PeLU nonlinearity~\cite{pelu_icmla17}. The reduction layer and DCF are trained by an efficient backprop formulation (see~\cite{atom_cvpr19} for details).  

\begin{figure}[!t]
\begin{center}
	\includegraphics[width=\linewidth]{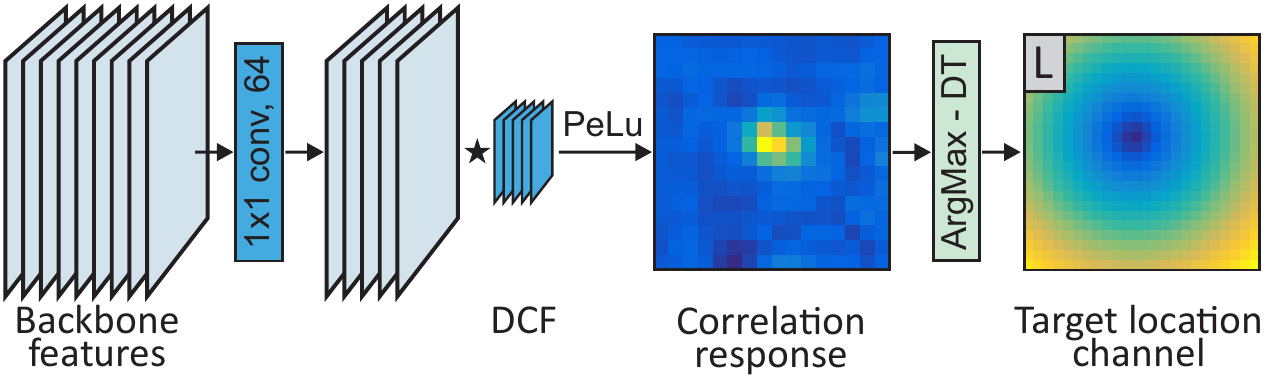}
\end{center}
   \caption{GEM -- the geometrically constrained Euclidean model -- reduces the backbone features dimensionality and correlates them with a DCF. 
   The target localisation channel ($\mathbf{L}$)
   is the distance transform to the maximum correlation response, representing the per-pixel confidence of target presence.} 
\label{fig:scm-scheme} 
\end{figure}

The maximum of the correlation response is considered as the most likely target position.
The D3S output (i.e., segmentation), however, requires specifying a belief of target presence at each pixel. 
Therefore, a target location channel is constructed by computing a (Euclidean) distance transform from the position of the maximum in the correlation map to the remaining pixels in the search region. 
The GEM pathway is shown in Figure~\ref{fig:scm-scheme}.

\subsection{Refinement pathway}  \label{sec:refinement}

The GIM and GEM pathways provide complementary information about the pixel-level target presence. 
GEM provides a robust, but rather inaccurate estimate of the target region, whereas the output channels from GIM show a greater detail, but are less discriminative (Figure~\ref{fig:intro_figure}). 
Furthermore, the individual outputs are low-resolution due to the backbone encoding. A refinement pathway is thus designed to combine the different information channels and upscale the solution into an accurate and detailed segmentation map. 

\begin{figure}[!t]
\begin{center}
	\includegraphics[width=\linewidth]{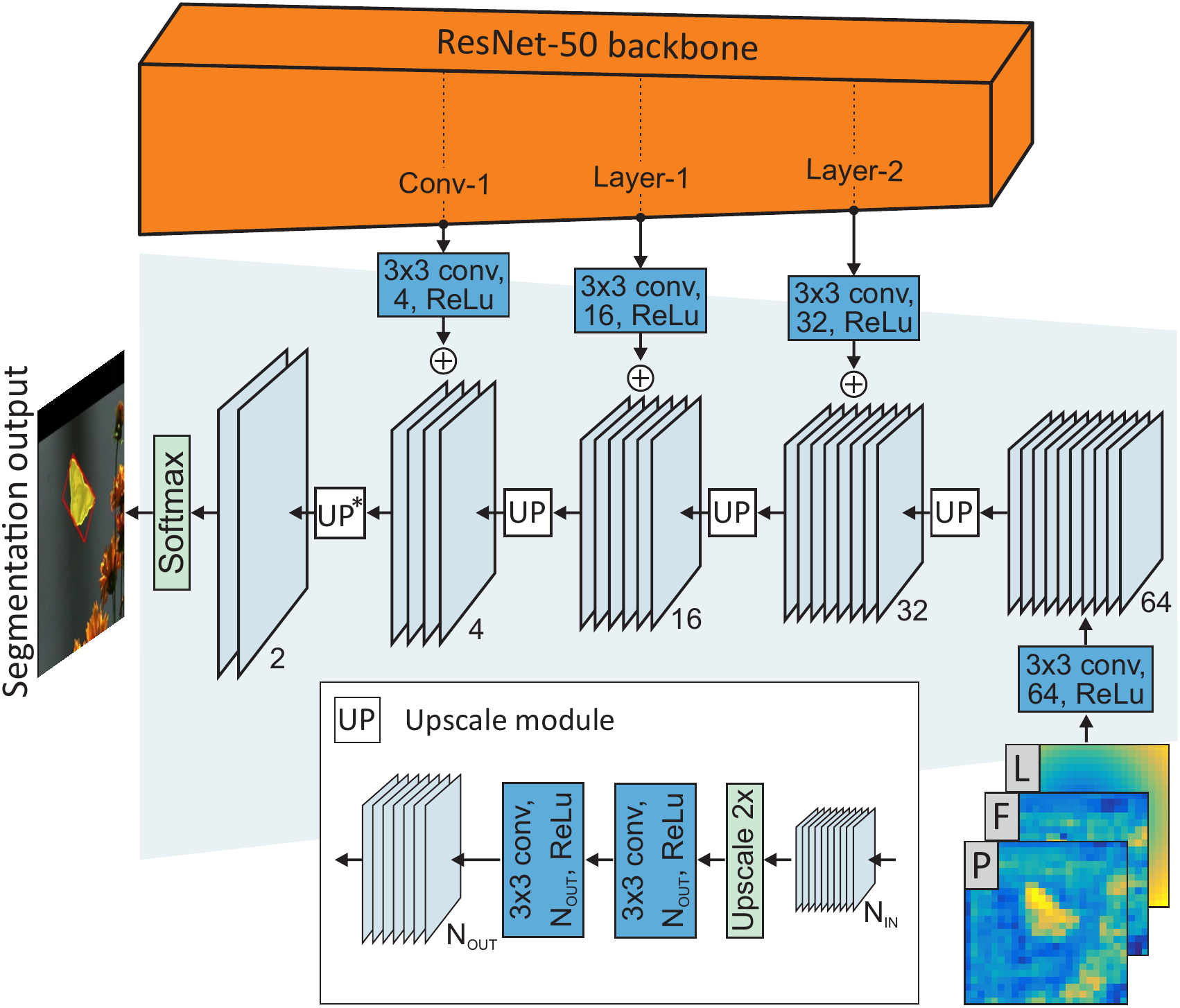}
\end{center}
   \caption{ The refinement pathway combines the GIM and GEM channels and gradually upscales  them by using adjusted features from the backbone. The UP$^*$ is a modified UP layer (see~the~
text).
   } 
\label{fig:refinement-architecture} 
\end{figure}

The refinement pathway takes the following inputs: the target location channel ($\mathbf{L}$) from GEM and the foreground similarity and posterior channels ($\mathbf{F}$ and $\mathbf{P}$) from the GIM. 
The channels are concatenated and processed by a $3 \times 3$ convolutional layer followed by a ReLU, resulting in a tensor of 64 channels. 
Three stages of upscaling akin to~\cite{unet_2015,refine_eccv16} are then applied to refine the details by considering the features in different layers computed in the backbone. 
An upscaling stage consists of doubling the resolution of the input channels, 
followed by two $3 \times 3$ convolution layers (each followed by a ReLU).
The resulting channels are summed with the adjusted features from the corresponding backbone layer. Specifically, the backbone features are adjusted for the upscaling task by a $3\times 3$ convolution layer, followed by a ReLU.
The last upscaling stage (which contains only resolution doubling, followed by a single $3\times3$ convolution layer) is followed by a softmax to produce the final segmentation probability map.
The refinement pathway is shown in Figure~\ref{fig:refinement-architecture}.

\section{Discriminative Segmentation Tracker}  \label{sec:tracker}  

This section outlines application of the discriminative segmentation network from Section~\ref{sec:methods} to online general object tracking. 
Given a single supervised training example from the first frame, the network produces target segmentation masks in all the remaining frames. 
However, some applications and most tracking benchmarks require target location represented by a bounding box. For most benchmarks, the bounding box is trivially obtained by fitting an axis-aligned bounding box that tightly fits a segmentation mask. 
However, for the benchmark requiring a rotated bounding box, we propose a simple fitting procedure in Section~\ref{sec:bbox-fitting}. The tracking steps are outlined in Section~\ref{sec:diset-tracking}. 

\subsection{Bounding box fitting module}  \label{sec:bbox-fitting}

The segmentation probability map from the discriminative segmentation network (Section~\ref{sec:methods}) is thresholded at $0.5$ probability to yield a binary segmentation mask.
Only the largest connected component within the mask is kept and an ellipse is fitted to its outline by least squares~\cite{Fitzgibbon_BMVC95}. The ellipse center, major and minor axis make up an initial estimate of the rotated bounding box. 
This is typically the most liberal solution with oversized rectangles, preferring most of the target pixels lying within its area, but accounts poorly for the presence of the background pixels within the region. 
We therefore further reduce the rectangle sides in direction of the major axes by optimizing the following modified overlap cost function $IoU^{\mathrm{MOD}}$ between the predicted segmentation mask and fitted rectangle using a coordinate descent:
\begin{equation}  \label{eq:iou_mod}
    IoU^{\mathrm{MOD}} = \frac{N^{+}_{\mathrm{IN}}}{\alpha N^{-}_{\mathrm{IN}} + N^{+}_{\mathrm{IN}} + N^{+}_{\mathrm{OUT}}},
\end{equation} 
where $N^{+}_{\mathrm{IN}}$ and $N^{+}_{\mathrm{OUT}}$ denote the number of foreground pixels within and outside the rectangle, respectively, and $N^{-}_{\mathrm{IN}}$ denotes the number of background pixels within the rectangle. 
The scalar $\alpha$ controls the contribution of $N^{-}_{\mathrm{IN}}$. The bounding box fitting method is very fast and takes on average only 2ms.

\subsection{Tracking with D3S}  \label{sec:diset-tracking}
 
\textbf{Initialization.} D3S is initialized on the first frame using the ground truth target location. 
The GEM and GIM initialization details depend on whether the target ground truth is presented by a bounding box or a segmentation mask. 
If a ground truth bounding box is available, the GEM follows the initialization procedure proposed in~\cite{atom_cvpr19}, which involves training both the dimensionality reduction network and the DCF by backprop on the first frame by considering the region four times the target size. 
On the other hand, if a segmentation mask is available, the ground truth target bounding box is first approximated by an axis-aligned rectangle encompassing the segmented target.

In case a segmentation mask is available, the GIM is initialized by extracting foreground samples from the target mask and background samples from the neighborhood four times the target size. 
However, if only a bounding box is available, an approximate ground truth segmentation mask is constructed first.
Foreground samples are extracted from within the bounding box, while the background samples are extracted from a four times larger neighborhood. A tracking iteration of D3S is then run on the initialization region to infer a \textit{proxi} ground truth segmentation mask. 
The final foreground and background samples are extracted from this mask. This process might be iterated a few times (akin to GrabCut~\cite{rother_grabcut2004}), however, we did not observe improvements and chose only a single iteration for initialization speed and simplicity. 
 
\textbf{Tracking.} During tracking, when a new frame arrives, a region four times the target size is extracted at previous target location. 
The region is processed by the discriminative segmentation network from Section~\ref{sec:methods} to produce the output segmentation mask. 
A rotated bounding box is fitted to the mask (Section~\ref{sec:bbox-fitting}) if required by the evaluation protocol. The DCF in the GEM is updated on the estimated target location following the backprop update procedure~\cite{atom_cvpr19}.

\section{Experiments}  \label{sec:experiments}

\subsection{Implementation details}  \label{sec:implementation}

The backbone network in D3S is composed of the first four layers of ResNet50, pre-trained on ImageNet for object classification. 
The backbone features are extracted from the target search region resized to $384 \times 384$ pixels. The background tradeoff parameter from (\ref{eq:iou_mod}) is set to $\alpha=0.25$ and the top $K=3$ similarities are used in GIM (\ref{eq:fg-similarity}). 
We verified in a preliminary analysis that performance is insensitive to exact values of these parameters, and we therefore keep the same values in all experiments. 

\noindent{\bf Network pre-training.} The GIM pathway and the refinement pathway are pre-trained on 3471 training segmentation sequences from  Youtube-VOS~\cite{yt_vos2018}. 
A training sample is constructed by uniformly sampling a pair of images and the corresponding segmentation masks from the same sequence within a range of 50 frames. 
To increase the robustness to possibly inaccurate GEM localization, the target location channel was constructed by perturbing ground truth locations uniformly from $[-\frac{1}{8}\sigma, \frac{1}{8}\sigma]$, where $\sigma$ is target size.
The network was trained by 64 image pairs batches for 40 epochs with 1000 iterations per epoch using the ADAM optimizer~\cite{adam_2015} with learning rate set to $10^{-3}$ and with 0.2 decay every 15 epochs. 
The training loss was a crossentropy between the predicted and ground truth segmentation mask. The training takes 20 hours on a single GPU.

\noindent{\bf Speed.} A Pytorch implementation of D3S runs at 25fps on a single NVidia GTX 1080 GPU, while 1.3s is required for loading the network to GPU and initialization.

\subsection{Evaluation on Tracking Datasets}  \label{sec:experiments-tracking}

\begin{table*}[!th]
\begin{center}
\begin{tabular}{l r r r r r r r r r}
\hline
 & \multicolumn{1}{c}{D3S} & \multicolumn{1}{c}{SPM} & \multicolumn{1}{c}{SiamMask} & \multicolumn{1}{c}{ATOM} & \multicolumn{1}{c}{ASRCF} & \multicolumn{1}{c}{SiamRPN} & \multicolumn{1}{c}{CSRDCF} & \multicolumn{1}{c}{CCOT} & \multicolumn{1}{c}{TCNN} \\
\hline
EAO~$\uparrow$  & \first{0.493} & \second{0.434} & \third{0.433} & 0.430 & 0.391 & 0.344 & 0.338 & 0.331 & 0.325 \\
Acc.~$\uparrow$ & \first{0.66}\phantom{0}  & \third{0.62}\phantom{0}  & \second{0.64}\phantom{0}  & 0.61\phantom{0}  & 0.56\phantom{0}  & 0.56\phantom{0}  & 0.51\phantom{0}  & 0.54\phantom{0}  & 0.55\phantom{0} \\
Rob.~$\downarrow$ & \first{0.131} & 0.210 & 0.214 & \second{0.180} & \third{0.187} & 0.302 & 0.238 & 0.238 & 0.268 \\
\hline
\end{tabular}
\end{center}
\caption{VOT2016 -- comparison with state-of-the-art trackers.} 
\label{tab:VOT2016}
\end{table*}

\begin{table*}[!th]
\begin{center}
\begin{tabular}{l r r r r r r r r}
\hline  
 & \multicolumn{1}{c}{D3S} & \multicolumn{1}{c}{SiamRPN++} & \multicolumn{1}{c}{ATOM} & \multicolumn{1}{c}{LADCF} & \multicolumn{1}{c}{DaSiamRPN} & \multicolumn{1}{c}{SiamMask} & \multicolumn{1}{c}{SPM} & \multicolumn{1}{c}{ASRCF} \\
\hline
EAO~$\uparrow$  & \first{0.489} & \second{0.414} & \third{0.401} & 0.389 & 0.383 & 0.380 & 0.338 & 0.328 \\
Acc.~$\uparrow$ & \first{0.64}\phantom{0}  & \third{0.60}\phantom{0}  & 0.59\phantom{0}  & 0.51\phantom{0}  & 0.59\phantom{0}  & \second{0.61}\phantom{0}  & 0.58\phantom{0}  & 0.49\phantom{0} \\
Rob.~$\downarrow$ & \first{0.150} & 0.234 & \third{0.204} & \second{0.159} & 0.276 & 0.276 & 0.300 & 0.234 \\
\hline
\end{tabular}
\end{center}
\caption{VOT2018 -- comparison with state-of-the-art trackers.} 
\label{tab:VOT2018}
\end{table*}

\begin{table*}[!th]
\begin{center}
\begin{tabular}{c c c c c c c c c }
\hline  
 & D3S & ATOM & SiamMask & SiamFCv2 & SiamFC & GOTURN & CCOT & MDNet \\
\hline
AO~$\uparrow$ & \first{59.7} & \second{55.6} & \third{51.4}  & 37.4 & 34.8 & 34.2 & 32.5 & 29.9 \\
SR$_{0.75}$~$\uparrow$ & \first{46.2} & \second{40.2} & \third{36.6} & 14.4 & 9.8 & 12.4 & 10.7 & \phantom{0}9.9 \\
SR$_{0.5}$~$\uparrow$ & \first{67.6} & \second{63.5} & \third{58.7} & 40.4 & 35.3 & 37.5 & 32.8 & 30.3 \\
\hline
\end{tabular}
\end{center}
\caption{GOT-10k test set -- comparison with state-of-the-art trackers .} 
\label{tab:got10k}
\end{table*}

\begin{table*}[!th]
\begin{center}
\begin{tabular}{l r r r r r r r r}
\hline  
 & D3S & SiamRPN++ & SiamMask & ATOM & MDNet & CFNet & SiamFC & ECO \\
\hline
AUC~$\uparrow$ & \second{72.8} & \first{73.3} & \third{72.5} & 70.3 & 60.6 & 57.8 & 57.1 & 55.4 \\
Prec.~$\uparrow$ & \second{66.4} & \first{69.4} & \second{66.4} & \third{64.8} & 56.5 & 53.3 & 53.3 & 49.2 \\
Prec.$_{\mathrm{N}}$~$\uparrow$ & 76.8 & \first{80.0} & \second{77.8} & \third{77.1} & 70.5 & 65.4 & 66.3 & 61.8 \\
\hline
\end{tabular}
\end{center}
\caption{TrackingNet test set -- comparison with state-of-the-art trackers.} 
\label{tab:trackingnet}
\end{table*}
 
D3S was evaluated on four major short-term tracking datasets:  VOT2016~\cite{kristan_vot2016}, VOT2018~\cite{kristan_vot2018}, GOT-10k~\cite{got10k} and TrackingNet~\cite{muller_trackingnet}. In the following we discuss the results obtained on each of the datasets.

{\bf VOT2016} and {\bf VOT2018} datasets each consist of 60 sequences. Targets are annotated by rotated rectangles to enable a more thorough localization accuracy evaluation compared to the related datasets. 
The standard VOT evaluation protocol~\cite{kristan_vot_tpami2016} is used in which the tracker is reset upon tracking failure. Performance is measured by accuracy (average overlap over successfully  tracked  frames),  robustness (failure  rate) and the EAO (expected average overlap), which is a principled combination of the former two measures~\cite{kristan_vot2015}.
 
The following state-of-the-art (sota) trackers are considered on {\bf VOT2016}: the VOT2016 top performers CCOT~\cite{danelljan_eccv2016_ccot} and TCNN~\cite{tcnn_2016}, a sota segmentation-based discriminative correlation filter CSR-DCF~\cite{Lukezic_CVPR_2017}, and most recently published sota deep trackers SiamRPN~\cite{siamrpn_cvpr2018}, SPM~\cite{spm_cvpr2019}, ASRCF~\cite{asrcf_cvpr2019}, SiamMask~\cite{siammask_cvpr19} and ATOM~\cite{atom_cvpr19}. 

Results reported in Table~\ref{tab:VOT2016} show that D3S outperforms all tested trackers on all three measures by a large margin. 
In EAO measure, D3S outperforms the top sota tracker SPM by 14\%, and simultaneously outperforms the top robust sota ATOM by 25\% in robustness. 
The top sota performer in accuracy is the segmentation-based tracker SiamMask. D3S outperforms this tracker by over 3\% in accuracy and approximately by 50\% in robustness.

The VOT2016 dataset contains per-frame target segmentation masks which can be used to evaluate segmentation performance on the small and challenging targets present. 
We have thus compared D3S with the most recent segmentation tracker SiamMask by computing the average IoU between the ground truth and predicted segmentation masks during periods of successful tracks (i.e., segmentation accuracy). 
D3S achieves a 0.66 average IoU, while SiamMask  IoU is 0.63. A nearly 5\% improvement speaks of a considerable accuracy of the D3S segmentation mask prediction. 

On the {\bf VOT2018} dataset, D3S is compared with the following sota trackers: the top VOT2018 performer LADCF~\cite{ladcf_tip2019} and the most recent sota trackers DaSiamRPN~\cite{dasiamrpn_eccv2018}, SiamRPN++~\cite{siamrpn_cvpr2019}, ATOM~\cite{atom_cvpr19}, SPM~\cite{spm_cvpr2019}, ASRCF~\cite{asrcf_cvpr2019} and SiamMask~\cite{siammask_cvpr19}. 
Results are reported in Table~\ref{tab:VOT2018}. Again, D3S outperforms all sota trackers in all measures. 
The top sota trackers in EAO, accuracy and robustness are SiamRPN++, SiamMask and LADCF, respectively. 
D3S outperforms the SiamRPN++ in EAO by 18\%, SiamMask in accuracy by over 5\% and LADCF by over 6\% in robustness. 
Note that SiamMask is a segmentation tracker, which explains the top accuracy among sota. D3S outperforms this tracker by over 45\% in robustness, which is attributed to the discriminative formulation within the single-pass segmentation mask computation.
 
{\bf GOT-10k} is a recent large-scale high-diversity dataset consisting of 10k video sequences with targets annotated by axis-aligned bounding boxes. The trackers are evaluated on 180 test sequences with 84 different object classes and 32 motion patterns, while the rest of the sequences form a training set. 
A tracker is initialized on the first frame and let to track to the end of the sequence. 
Trackers are ranked according to the average overlap, but success rates (SR$_{0.5}$ and SR$_{0.75}$) are reported at two overlap thresholds 0.5 and 0.75, respectively, for detailed analysis\footnote{Success rate denotes percentage of frames where predicted region overlaps with the ground-truth region more than the threshold.}. The following top-performing sota trackers are used in comparison~\cite{got10k}: SiamFCv2~\cite{Valmadre_2017_CVPR}, SiamFC~\cite{siamfc_eccv16}, GOTURN~\cite{goturn_eccv2016}, CCOT~\cite{danelljan_eccv2016_ccot}, MDNet~\cite{mdnet_cvpr2016} and the most-recent ATOM~\cite{atom_cvpr19} and SiamMask~\cite{siammask_cvpr19}.
We emphasize that D3S is not fine-tuned on the training set, while some of the top-performing sota trackers we use in comparison do utilize the GOT-10k training set.
Results on GOT-10k are reported in Table~\ref{tab:got10k}.
D3S outperforms all top-performing sota by a large margin in all performance measures, and achieves approximately 60\% boost in average overlap compared to the SiamFCv2, which is a top-performer on~\cite{got10k} benchmark. It also outperforms the most recent ATOM and SiamMask trackers by over 7\% and over 15\% in average overlap, respectively.
This demonstrates considerable generalization ability over a diverse set of target types.

{\bf TrackingNet} is another large-scale dataset for training and testing trackers.  
The training set consists of over 30k video sequences, while the testing set contains 511 sequences.  A tracker is initialized on the first frame and let to track to the end of the sequence. Trackers are ranked according to the area under the success rate curve (AUC), precision (Prec.) and normalized precision (Prec.$_N$). 
The reader is referred to~\cite{muller_trackingnet} for further details about the performance measures. 
The performance of D3S is compared  with the top-performing sota trackers according to~\cite{muller_trackingnet}: ECO~\cite{DanelljanCVPR2017}, SiamFC~\cite{siamfc_eccv16}, CFNet~\cite{Valmadre_2017_CVPR}, MDNet~\cite{mdnet_cvpr2016} and most recent sota trackers ATOM~\cite{atom_cvpr19}, SiamMask~\cite{siammask_cvpr19} and SiamRPN++~\cite{siamrpn_cvpr2019}.
D3S significantly outperforms the sota reported in~\cite{muller_trackingnet} and is on par with SiamRPN++, SiamMask and ATOM. 
Note that D3S is trained only on 3471 sequences from YouTube-VOS~\cite{yt_vos2018}, while both, ATOM and SiamRPN++ are finetuned on much larger datasets (31k, and over 380k sequences, respectively), which include the TrackingNet training set. 
This further supports a considerable generalization capability of D3S, which is primarily trained for segmentation, not tracking.

\subsection{Ablation Study}  \label{sec:ablation}

\begin{table*}[!t]
\begin{center}
\begin{tabular}{l l l l l l l | l l }
 & \multicolumn{1}{c}{D3S} & \multicolumn{1}{c}{$\mathrm{\bar F}$} & \multicolumn{1}{c}{$\mathrm{\bar U}$} & \multicolumn{1}{c}{$\mathrm{\bar P}$} & \multicolumn{1}{c}{$\mathrm{\bar F \bar P}$} & \multicolumn{1}{c}{$\mathrm{\bar L}$} & \multicolumn{1}{c}{$\mathrm{MA}$} & \multicolumn{1}{c}{$\mathrm{MM}$} \\
\hline
EAO  & 0.489 & 0.467 & 0.452 & 0.423 & 0.357 & 0.251 & 0.444 & 0.398 \\
Acc. & 0.64  & 0.65  & 0.63  & 0.60  & 0.55  & 0.60  & 0.60 & 0.55 \\
Rob. & 0.150 & 0.187 & 0.173 & 0.211 & 0.234 & 0.567 & 0.160 & 0.173 \\
\hline
\end{tabular}
\end{center}
\caption{VOT2018 -- ablation study.
Removing: the GIM foreground similarity channel ({$\mathrm{\bar F}$}),
the GIM foreground probability channel ({$\mathrm{\bar P}$}),
both GIM channels ({$\mathrm{\bar F \bar P}$}) and
the GEM channel ({$\mathrm{\bar L}$}).
The DCF in GEM is updated from its own position estimation rather than position estimated by D3S ({$\mathrm{\bar U}$}).
D3S with a minimal area rotated bounding box 
($\mathrm{MA}$) and a min-max axis-aligned bounding box ($\mathrm{MM}$).
} 
\label{tab:ablation}
\vspace{-0.25cm}
\end{table*}

An ablation study was performed on VOT2018 using the reset-based protocol~\cite{kristan_vot_tpami2016} to expose the contributions of different components of the D3S architecture. 
The following variations of D3S were created: 
(i) D3S without the GIM foreground similarity channel $\mathbf{F}$ (D3S$^{\bar F}$); 
(ii) D3S without the GIM target posterior channel $\mathbf{P}$ (D3S$^{\mathrm{\bar P}}$) ; 
(iii) D3S with only the GEM output channel and without GIM channels $\mathbf{F}$ and $\mathbf{P}$ (D3S$^\mathrm{\bar F \bar P}$); 
(iv)  D3S without the GEM output channel $\mathbf{L}$ (D3S$^{\bar L}$); 
(v) D3S in which the DCF is not updated from the position estimated by D3S, but rather from the position estimated by the DCF in GEM (D3S$^{\bar U}$).
Two additional D3S versions with different bounding box fitting methods were included: a minimal area rotated bounding box that contains all foreground pixels (D3S$^{\mathrm{MA}}$) and a min-max axis-aligned bounding box (D3S$^\mathrm{MM}$). All variations were re-trained on the same dataset as the original D3S. 

Results of the ablation study are presented in Table~\ref{tab:ablation}. 
Removal of the foreground similarity channel from GIM (D3S$^{\mathrm{\bar F}}$) causes a 4.5\% performance drop, while removal of the target posterior channel (D3S$^{\mathrm{\bar P}}$) reduces the performance by 13.5\%. 
The accuracy of both variants is comparable to the original D3S, while the number of failures increases. 
In conclusion, each, foreground similarity and posterior channel individually contribute to robust target localization.

Removal of the entire GIM module i.e., $\mathbf{F}$ and $\mathbf{P}$ (D3S$^{\mathrm{\bar F \bar P}}$) reduces the overall tracking performance by 27\%. 
The accuracy drops by 14\%, while the number of failures increases by 56\%. 
This speaks of crucial importance of the GIM module for accurate segmentation as well as tracking robustness.

Removal of the GEM module (D3S$^{\mathrm{\bar L}}$) reduces the tracking performance by nearly 50\%. This is primarily due to significant reduction of the robustness -- the number of failures increases by over 270\%. 
Thus the GEM module is crucial for robust target selection in the segmentation process.
 
Finally, updating the DCF in GEM module by its own estimated position rather than the position estimated by the final segmentation (D3S$^{\mathrm{\bar U}}$) reduces the overall performance by 7.5\%, primarily at a cost of significant increase in the number of failures (over 15\%). 
Thus, accurate target position estimation from D3S crucially affects the learning of the DCF in GEM and consequently the overall tracking performance.

Replacing the proposed bounding box fitting method (Section~\ref{sec:bbox-fitting}) with the minimal area rotated bounding box (D3S$^{\mathrm{MA}}$) results in a 9\% reduction in EAO and a 6\% reduction in accuracy. 
This is still a state-of-the-art result, which means that the D3S performance boost can be primarily attributed to the segmentation mask quality. The min-max bounding box fitting method (D3S$^{\mathrm{MM}}$) leads to a 19\% EAO and 14\% accuracy reduction. 
Thus D3S does benefit from the rotated bounding box estimation.

\subsection{Evaluation on Segmentation Datasets}  \label{sec:experiments-segmentation}

Segmentation capabilities of D3S were analyzed on two popular video object segmentation benchmarks DAVIS16~\cite{davis16} and DAVIS17~\cite{davis17}.
Under the DAVIS protocol, the segmentation algorithm is initialized on the first frame by a segmentation mask. The algorithm is then required to output the segmentation mask for all the remaining frames in the video. Performance is evaluated by two measures averaged over the sequences: mean Jaccard index ($\mathcal{J_{M}}$) and mean F-measure ($\mathcal{F_{M}}$).
Jaccard index represents a per-pixel intersection over union between the ground-truth and the predicted segmentation mask. The F-measure is a harmonic mean of precision and recall calculated between the contours extracted from the ground-truth and the predicted segmentation masks. 
For further details on these performance measures, the reader is referred to~\cite{davis16,martin_pami2004}.

D3S is compared to several sota video object segmentation methods specialized to the DAVIS challenge setup: OSVOS~\cite{osvos_cvpr2017}, OnAVOS~\cite{onavos_bmvc2017}, OSMN~\cite{osmn_cvpr2018}, FAVOS~\cite{favos_cvpr2018}, VM~\cite{videomatch_eccv2018} and PML~\cite{blazingly_fast_cvpr18}. 
In addition, we include the most recent segmentation-based tracker SiamMask~\cite{siammask_cvpr19}, which is the only published method that performs well on both, short-term tracking as well as on video object segmentation benchmarks.

Results are shown in Table~\ref{tab:davis}. D3S performs on par with most of the video  object segmentation top performers on DAVIS. 
Compared to top performer on DAVIS2016, the performance of D3S is 12\% and 14\% lower in the average Jaccard index and the F-measure, respectively. 
On DAVIS2017 this difference is even smaller -- a 6\% drop in Jaccard index and 8\% drop in F-measure compared to the top-performer OnAVOS. 
This is quite remarkable, considering that D3S is 200 times faster. Furthermore, D3S delivers a comparable segmentation accuracy as pure segmentation methods ASMN and PML, while being orders of magnitude faster and achieving a near-realtime video object segmentation, which is particularly important for many video editing applications.
 
\begin{figure}[!t]
\begin{center}
	\includegraphics[width=\linewidth]{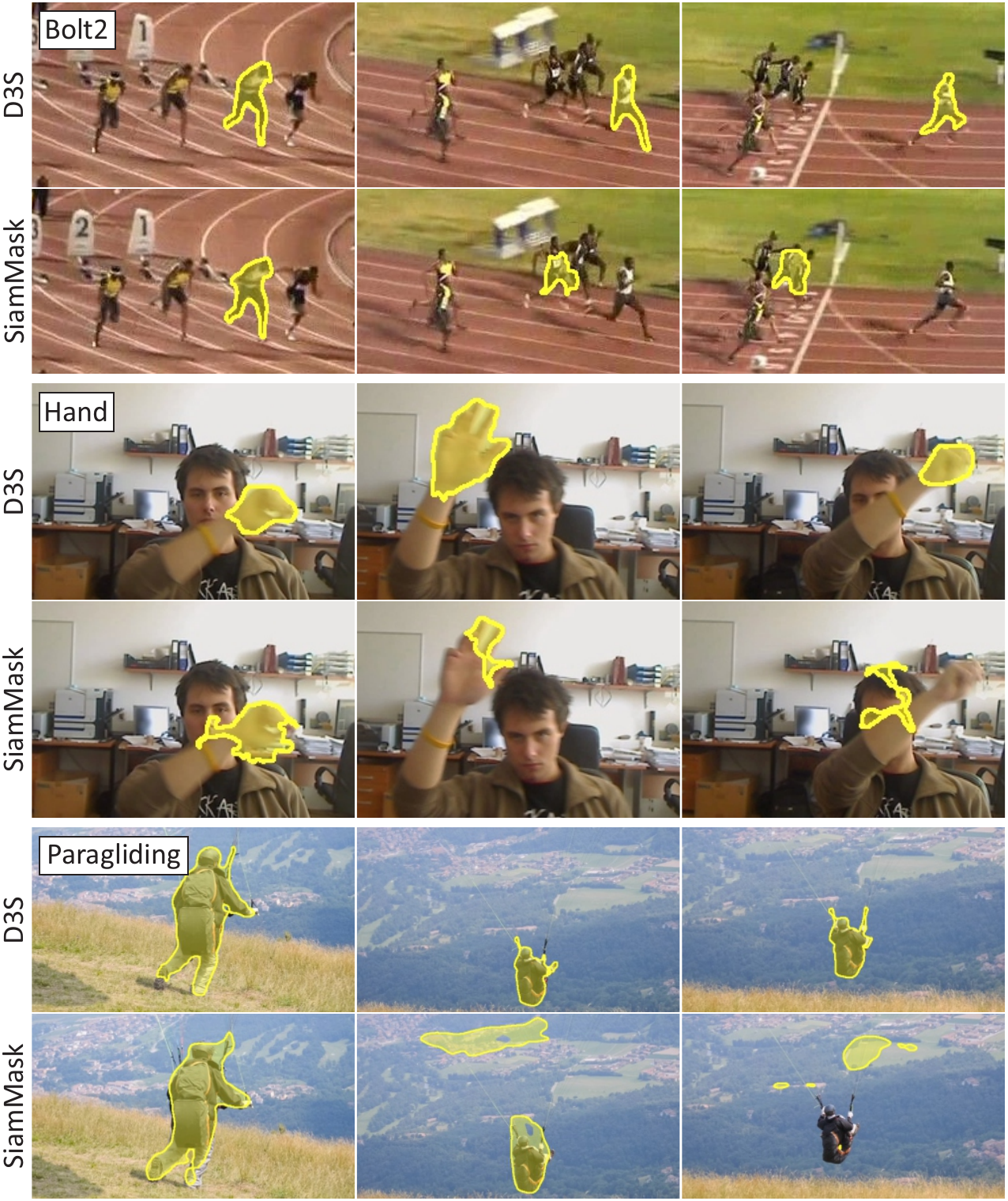}
\end{center}
    \vspace{-0.1cm}
   \caption{D3S vs. SiamMask segmentation quality. 
   \textit{Bolt2}: SiamMask drifts to a similar object, D3S leverages the discriminative learning in GEM to robustly track the selected target.
   \textit{Hand}: the rigid template in SiamMask fails on a deforming target, the GIM model in D3S successfully tracks despite a significant deformation. 
   \textit{Paragliding}: clutter causes drift and failure of SiamMask while in D3S, the combination of the GIM and GEM models leads to accurate and robust segmentation. } 
\label{fig:qualitative-segmentation-comparison}
\vspace{-0.25cm}
\end{figure}

D3S also outperforms the only tracking and segmentation method SiamMask with respect to all measures. On average the segmentation is improved by over 5\% in the Jaccard index and the contour accuracy-based F-measure. See Figure~\ref{fig:qualitative-segmentation-comparison} for further qualitative comparison of D3S and SiamMask on challenging targets.

\begin{table}[t]
\begin{center}
\scalebox{0.9}{
\begin{tabular}{l c c c c r }
\hline  
 & $\mathcal{J_{M}}^{16}$ & $\mathcal{F_{M}}^{16}$ & $\mathcal{J_{M}}^{17}$ & $\mathcal{F_{M}}^{17}$ & FPS \\ 
\hline
D3S                              & 75.4 & 72.6 & 57.8 & 63.8 & 25.0  \\
SiamMask~\cite{siammask_cvpr19}  & 71.7 & 67.8 & 54.3 & 58.5 & 55.0  \\
OnAVOS~\cite{onavos_bmvc2017}        & 86.1 & 84.9 & 61.6 & 69.1 & 0.1 \\
FAVOS~\cite{favos_cvpr2018}      & 82.4 & 79.5 & 54.6 & 61.8 & 0.8 \\
VM~\cite{videomatch_eccv2018}    & 81.0 &   -  & 56.6 &   -  & 3.1 \\
OSVOS~\cite{osvos_cvpr2017}      & 79.8 & 80.6 & 56.6 & 63.9 & 0.1 \\
PML~\cite{blazingly_fast_cvpr18} & 75.5 & 79.3 &   -  &   -  & 3.6 \\
OSMN~\cite{osmn_cvpr2018}        & 74.0 & 72.9 & 52.5 & 57.1 & 8.0   \\
\hline
\end{tabular}
}
\end{center}
\caption{State-of-the-art comparison on the DAVIS16 and DAVIS17 segmentation datasets. Average Jaccard index and F-measure are denoted as $\mathcal{J_{M}}^{16}$ and $\mathcal{F_{M}}^{16}$ on DAVIS16 dataset and $\mathcal{J_{M}}^{17}$ and $\mathcal{F_{M}}^{17}$ on DAVIS17 dataset, respectively.} 
\label{tab:davis}
\vspace{-0.25cm}
\end{table}

\section{Conclusion}  \label{sec:conclusion}

A deep single-shot discriminative segmentation tracker -- D3S -- was introduced. The tracker leverages two  models from the extremes of the spectrum: a geometrically invariant model and a geometrically restricted Euclidean model. 
The two models localize the target in parallel pathways and complement each other to achieve high segmentation accuracy of deformable targets and robust discrimination of the target from distractors. 
The end-to-end trainable network architecture is the first single-shot pipeline with online adaptation that tightly connects discriminative tracking with accurate segmentation.

D3S outperforms state-of-the-art trackers on the VOT2016, VOT2018 and GOT-10k benchmarks and performs on par with top trackers on TrackingNet, regardless of the fact that some of the tested trackers were re-trained for specific datasets. 
In contrast, D3S was trained once on Youtube-VOS (for segmentation only) and the same version was used in all benchmarks.
Tests on DAVIS16 and DAVIS17 segmentation benchmarks show performance close to top segmentation methods while running up to 200$\times$ faster, close to real-time. 
D3S significantly outperforms recent top segmentation tracker SiamMask on all bechmarks in all metrics and contributes towards narrowing the gap between two, currently separate, domains of short-term tracking and video object segmentation, thus blurring the boundary between the two.

{
\footnotesize
\paragraph{Acknowledgements.}
This work was supported by Slovenian research agency program P2-0214 and projects J2-8175 and J2-9433. 
A. Lukežič was financed by the Young researcher program of the ARRS.
J. Matas was supported by the Czech Science Foundation grant GA18-05360S.
}

{\small
\bibliographystyle{ieee_fullname}
\bibliography{bib}
}

\newpage
\section*{Qualitative examples}  \label{sec:qualitative}

\begin{figure*}
\centering
\includegraphics[width=1\linewidth]{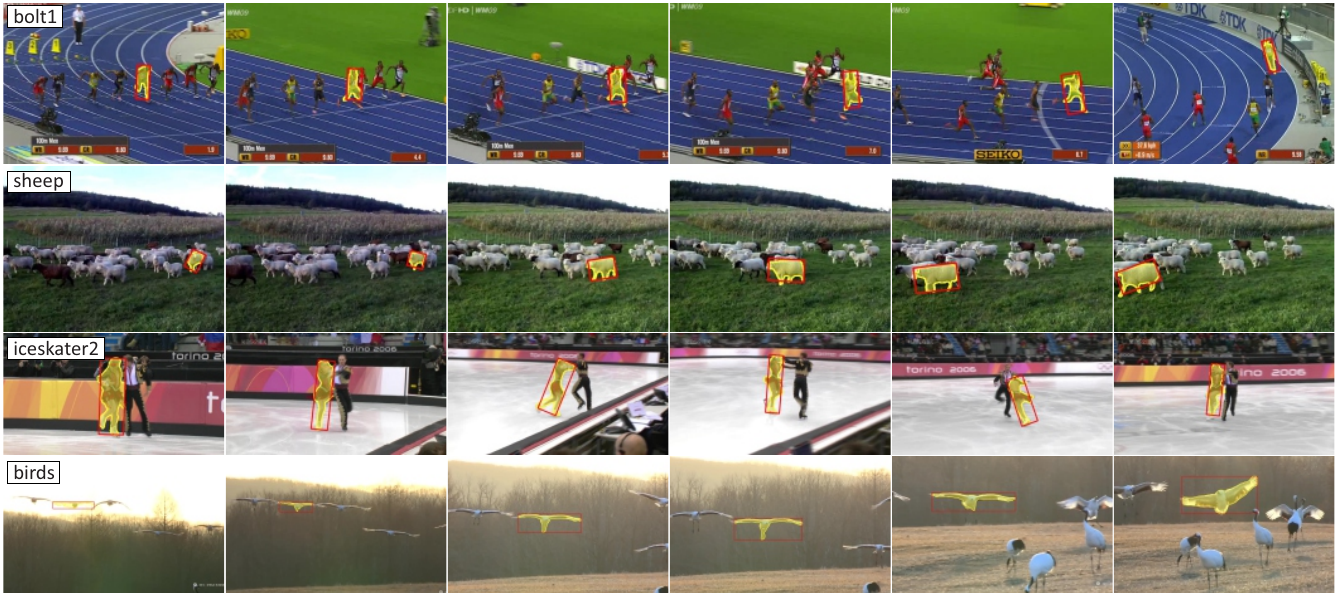}
\caption{Sequences with distractors (similar objects in the target vicinity). D3S segments the correct target even though a similar target is close (or even overlapping). These examples show discriminative power of the proposed tracker achieved by the discriminative GIM and GEM.}
\label{fig:similar}
\end{figure*}

\begin{figure*}[]
\centering
\includegraphics[width=1\linewidth]{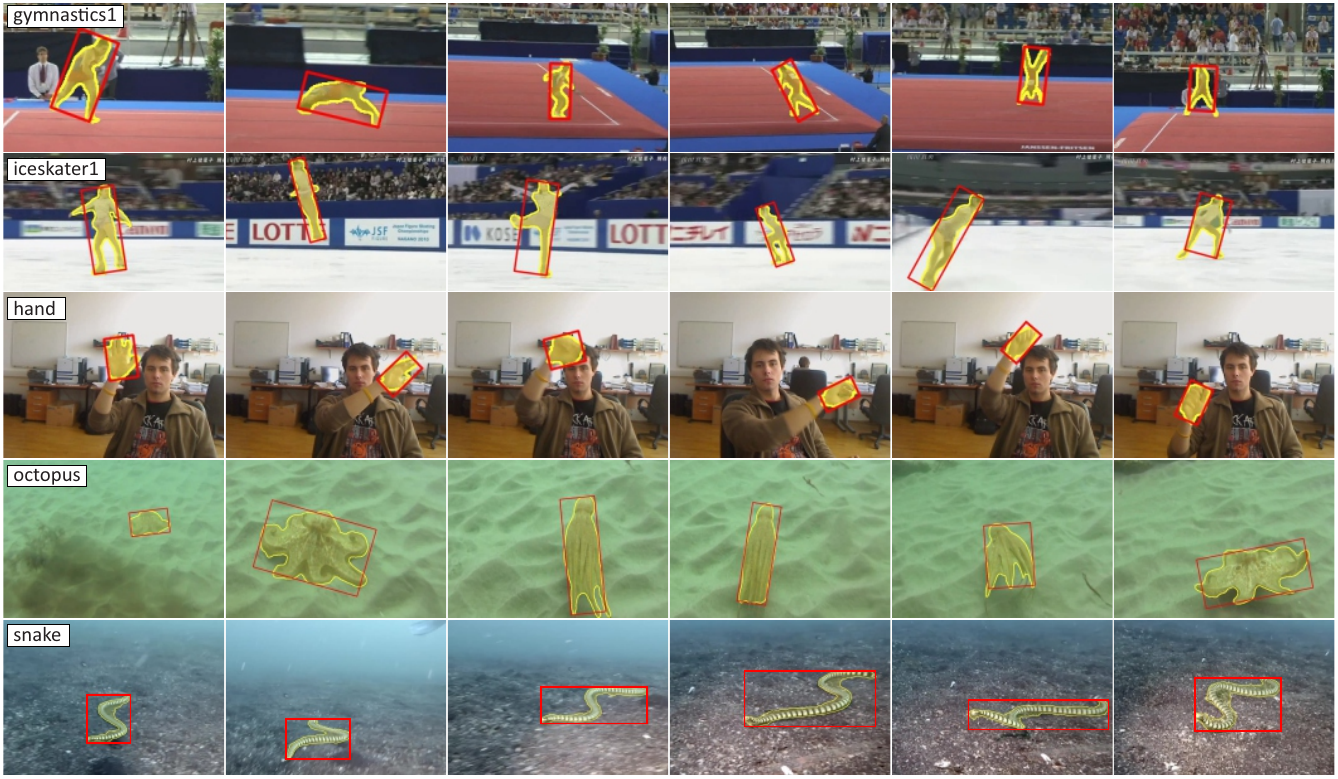}
\caption{Examples of appearance changes and deforming targets. The geometrically invariant model (GIM) successfully segments the target due to geometrically unrestricted representation even under target rotation ({\it gymnastics1}), articulated ({\it iceskater1} and {\it octopus}) or significantly change its shape ({\it hand} and {\it snake}).}
\label{fig:deformable}
\end{figure*}

\begin{figure*}
\centering
\includegraphics[width=1\linewidth]{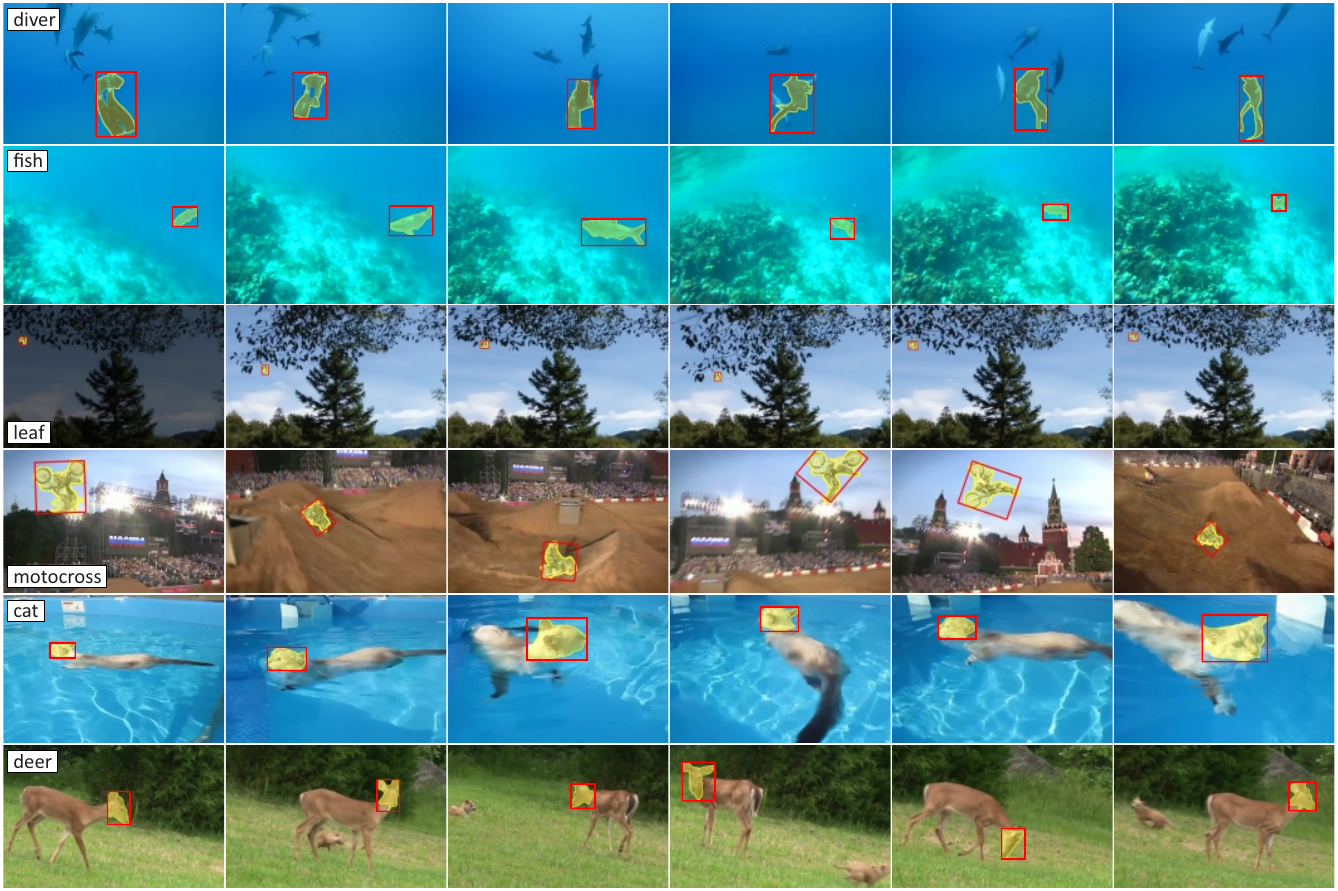}
\caption{Difficult examples to track and segment. 
Underwater video sequences {\it diver} and {\it fish} are challenging due to the low contrast between the target and background -- the D3S refinement pathway still produces an accurate segmentation. 
Small target in {\it leaf} sequence is successfully tracked and segmented due to the large search range (4-times of target size) and the discriminative architecture, even though several similar leaves are in the vicinity and all leaves undergo abrupt motion due to a high wind.
Target rotation and scale change in {\it motocross} sequence are successfully addressed by the geometrically invariant model (GIM).
A challenging scenario where only the head of the {\it cat} and {\it deer} is tracked. Foreground and background feature vectors in GIM and combination with GEM prevent segmenting the whole animal as the target.
}
\label{fig:difficult}
\end{figure*}

\begin{figure*}
\centering
\includegraphics[width=1\linewidth]{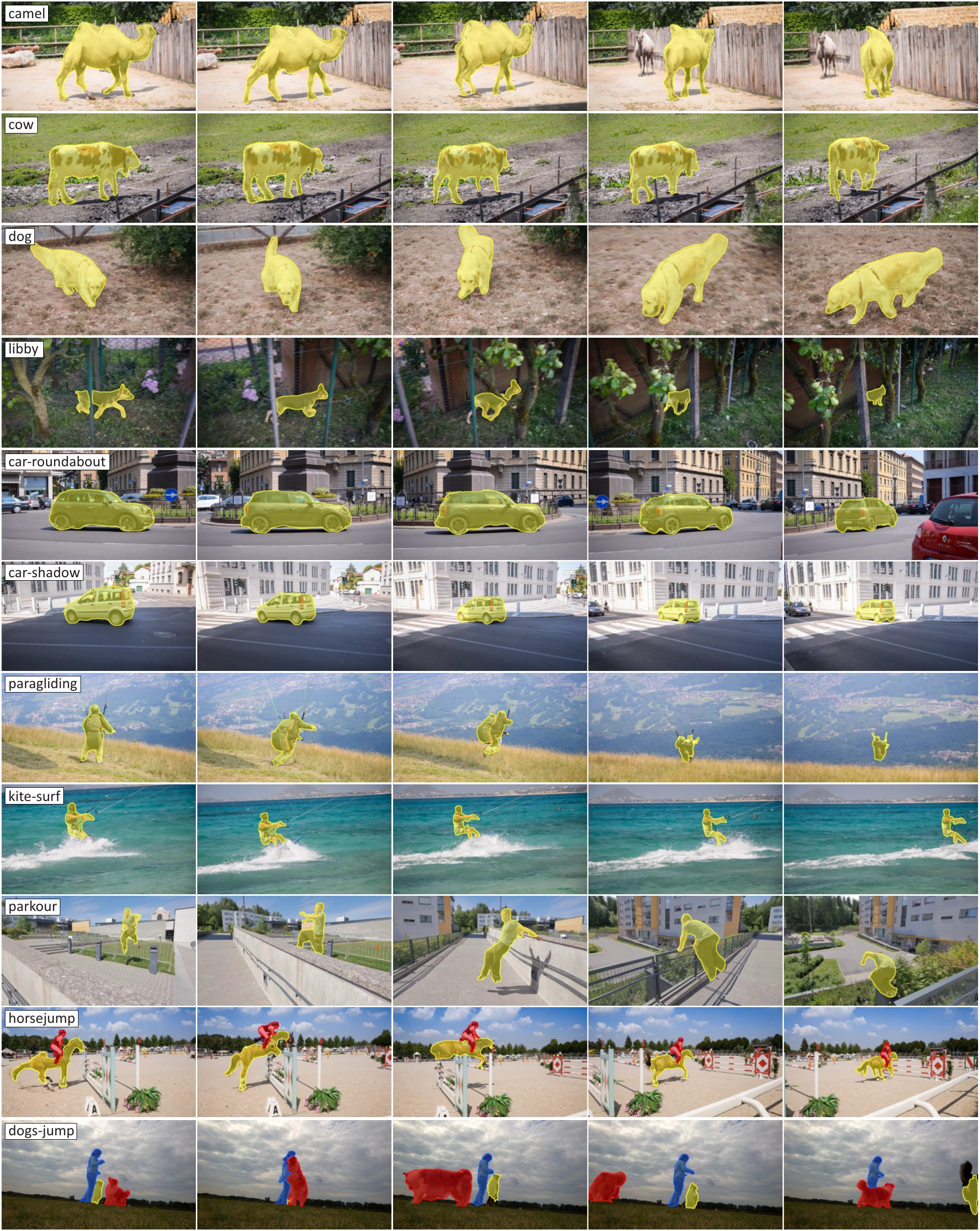}
\caption{Video object segmentation on DAVIS datasets. D3S produces a highly accurate segmentation in near real-time. In sequences with multiple objects the tracker was run independently on each target.}
\label{fig:davis}
\end{figure*}

We provide here additional qualitative examples of tracking and segmentation. 
Video sequences are collected from the VOT2016~\cite{kristan_vot2016}, GOT-10k~\cite{got10k} and DAVIS~\cite{davis16,davis17} datasets.
Output of the D3S is segmentation mask and it is visualized with yellow color. A bounding box is fitted to the predicted segmentation mask and shown in red.
Tracker reports binary segmentation mask for DAVIS, rotated bounding box for VOT sequences, while axis-aligned bounding box is required by the GOT-10k evaluation protocol. 
The following tracking and segmentation conditions are visualized:
\begin{itemize}
    \item Figure~\ref{fig:similar} demonstrates the discriminative power of D3S by visualizing tracking in presence of distractors, i.e., visually similar objects.
    \item Figure~\ref{fig:deformable} shows a remarkable segmentation accuracy and robustness of D3S on tracking of deformable objects and \textit{parts} of objects.
    \item Figure~\ref{fig:difficult} shows tracking in sequences we have identified as particularly challenging for the current state-of-the-art. It includes small objects and tracking parts of objects.
    \item Figure~\ref{fig:davis} shows (near real-time) video object segmentation results on DAVIS16~\cite{davis16} and DAVIS17~\cite{davis17} datasets.
\end{itemize}

\end{document}